\documentclass{article}

\usepackage{graphicx}
\usepackage{booktabs}
\usepackage{amsmath}
\usepackage{hyperref}
\usepackage{amsfonts}
\usepackage{makecell}
\usepackage[preprint]{corl_2026} 

\title{Guided Diffusion with Distilled Vision-Language Reliability for Aerial Navigation}

\author{
  Ivan Valuev\\
  \And
  Iana Zhura\\
  \And
  Valerii Serpiva\\
  \And 
  Didar Seyidov\\
  \And
  Dzmitry Tsetserukou}

\begin{document}
\maketitle


\begin{abstract}
Autonomous UAV navigation is conventionally solved by pipelines that separate perception,
mapping, and planning into distinct stages, which propagates errors, accumulates latency, and
requires environment-specific retuning. End-to-end generative models remove these interfaces
by mapping raw observations directly to trajectories, but inherit a subtle failure mode:
trained on clean data, they cannot recognise when an observation is unreliable, and treat
degraded regions such as glass, mirrors, and overexposed surfaces as valid evidence for
planning. We present a reliability-aware diffusion planner for 3D UAV navigation. It
conditions trajectory generation on the observation together with a scene-level reliability
heatmap that marks where perception cannot be trusted, produced by a lightweight network that
distils the open-vocabulary reasoning of a vision-language model within the real-time planning
budget. To generalise to unseen environments without retraining, we steer the denoising
process with a differentiable two-stage ESDF cost that treats physical obstacles from depth
and virtual obstacles from highly unreliable regions on equal footing. In simulation and on a
real quadrotor, our planner produces markedly safer trajectories than a state-of-the-art
diffusion baseline, reducing the obstacle-violation rate from $40.3\%$ to $9.6\%$ and raising
the mean reliability of traversed regions from $0.588$ to $0.925$. Ablating the reliability
term alone drops mean reliability from $0.898$ to $0.783$, confirming it as the decisive
component, while distillation runs the framework up to $2\times$ faster than the full
vision-language model.
\end{abstract}


\section{Introduction}

Autonomous aerial navigation in unstructured indoor environments remains a
challenging problem in robot learning. A micro aerial vehicle must perceive the
scene, infer a feasible collision-free path, and act in real time under tight
onboard compute and power budgets. Classical navigation stacks decompose this into
a chain of separate modules for perception, mapping, planning, and control. While
modular, these pipelines discard information at every hand-designed interface,
accumulate latency, and allow early errors to propagate and amplify
downstream, degrading reliability in cluttered real-world
settings~\cite{hart1968astar,karaman2011rrtstar,williams2017mppi,ratliff2009chomp}.

End-to-end learned policies mitigate these limitations by mapping raw observations
directly to trajectories within a single model, preserving the full information
content of the input. Diffusion models are well-suited to this setting. In contrast
to optimisation-based planners that return a single deterministic solution, they
learn a conditional distribution over feasible trajectories~\cite{zhura2025swarmdiffusion}, allowing the
policy to represent the multimodality inherent to navigation rather than collapse
it to a point estimate.

A further difficulty is that classical pipelines assume the constructed map
faithfully represents the environment. This assumption fails in the presence of
glass, mirrors, specular reflections, and saturated illumination, where depth
sensors return confident yet incorrect measurements that occupancy-based planners
cannot detect~\cite{oleynikova2017voxblox,zhou2020egoplanner,ames2019cbf}. Rather
than trusting depth uniformly, we estimate the reliability of the depth measurement
directly from the RGB image, identifying regions in which perception is likely to
be misleading. This reliability estimate is then used as a conditioning signal that
guides the diffusion sampling process away from unreliable regions, integrating
perceptual reliability into trajectory generation rather than treating it as a
separate post-hoc check.

Robust reliability estimation across diverse indoor conditions benefits from the
semantic priors of Vision-Language Models (VLMs), which can identify glass, mirrors,
and reflective surfaces from language-grounded visual
cues~\cite{radford2021clip,luddecke2022clipseg,sahu2025anytraverseA}. Their
computational cost, however, is incompatible with onboard inference budgets. We
therefore distil the VLM into a compact student model that reproduces its perceptual
reliability estimates at a fraction of the cost, and deploy only the student at
runtime.

In this work, we present an end-to-end diffusion-based framework for safe autonomous
UAV navigation indoors. The framework conditions trajectory generation jointly on
visual observations, geometric context, and an RGB-derived reliability map, and
enforces geometric feasibility through guided sampling on a three-dimensional ESDF.
Our main contributions can thus be summarized as follows:

\begin{itemize}

    \item \textbf{End-to-end perception and planning.}
    A single diffusion model that maps raw observations directly to distributions
    over three-dimensional trajectories, removing the error-compounding interfaces
    of modular pipelines.

    \item \textbf{RGB-based reliability detection with diffusion guidance.}
    Estimation of RGBD-sensor reliability from the RGB image, incorporated as a
    conditioning signal that steers the diffusion sampler away from regions where
    perception is unreliable.

    \item \textbf{Distillation of vision-language reliability reasoning.}
    A teacher-student scheme in which a VLM supervises reliability estimation during
    training and is distilled into a compact student capable of real-time inference.

    \item \textbf{Training-free obstacle avoidance.}
    A guided sampling procedure that enforces geometric collision constraints using
    3D ESDF gradients without environment-specific retraining, supporting
    generalisation to unseen indoor layouts.

\end{itemize}

\section{Related Work}
	
\subsection{Diffusion-Based Navigation}

Diffusion policies recast navigation as sampling from a learned distribution of
trajectories conditioned on raw observations, replacing explicit maps and hand-tuned
cost functions. Conditioning jointly on images, range data, goals, and recent states
supports mapless operation, with training biased toward traversable regions improving
path quality on physical robots~\cite{liang2024dtg}. The mechanism most relevant to our
method is guided sampling: treating the gradient of a differentiable cost as a surrogate
for the classifier in classifier guidance~\cite{dhariwal2021diffusion} steers a
pretrained denoiser toward feasible paths and transfers to new scenes without
retraining~\cite{zeng2025navidiffusor}. Related designs unify directed navigation and
exploration through goal masking~\cite{sridhar2023nomad} or recover actions from
language-conditioned imagined futures~\cite{du2023unipi}.

Three limitations of these methods motivate our design. First, the conditioning signal
is typically produced by a separate and considerably larger perception backbone, so the
deployed system carries the cost of an auxiliary model alongside the denoiser rather than
a single compact policy. Second, the iterative denoising chain is slow relative to flight
dynamics, and although warm-starting from partially noised samples~\cite{ma2025codig} or
staging noise levels across planning steps~\cite{ye2025radp} raises replanting rates,
real-time onboard operation remains constrained. Third, and most critically, the guidance
encodes only geometric feasibility. None of these methods assesses whether the conditioning
measurement can be trusted, so failures on glass, mirrors, or overexposed surfaces are
absent from the planning objective. Our approach addresses all three: it distills the
conditioning model into a compact onboard predictor, keeps the guided sampler lightweight
for fast replanning, and injects an explicit measurement-reliability term into the guidance
so that perceptually hazardous regions are penalized during denoising rather than ignored.

\subsection{Vision-Language Models for Navigation}

Vision-language models supply the open-vocabulary semantics that geometry alone can not
provide. Image-language embeddings have been grounded into spatial maps so that language
queries resolve to navigable locations without task-specific labels~\cite{huang2023vlmaps},
embodiment-agnostic foundation models learn transferable goal-reaching from large
heterogeneous datasets~\cite{shah2023vint,shah2022gnm}, and language models act as high-level
planners that parse instructions into subgoals~\cite{shah2023lmnav}. More recent
vision-language-action (VLA) models close the loop end to end, mapping observations and
instructions directly to motion with strong cross-task generalisation~\cite{zhang2024navid}.

These models are poorly matched to autonomous aerial navigation. Their parameter counts and
latency exceed what embedded flight hardware can sustain at control rate, and they presuppose
a human supplying language prompts throughout operation, whereas a UAV must navigate
autonomously. Like diffusion planners, they also reason about scene content rather than
sensing reliability. We retain the semantic competence of vision-language pretraining while
removing its runtime cost and prompt dependence by distilling it offline into a compact
reliability predictor.

\section{Methods}
\begin{figure}[t]
  \centering
  \includegraphics[width=0.85\linewidth]{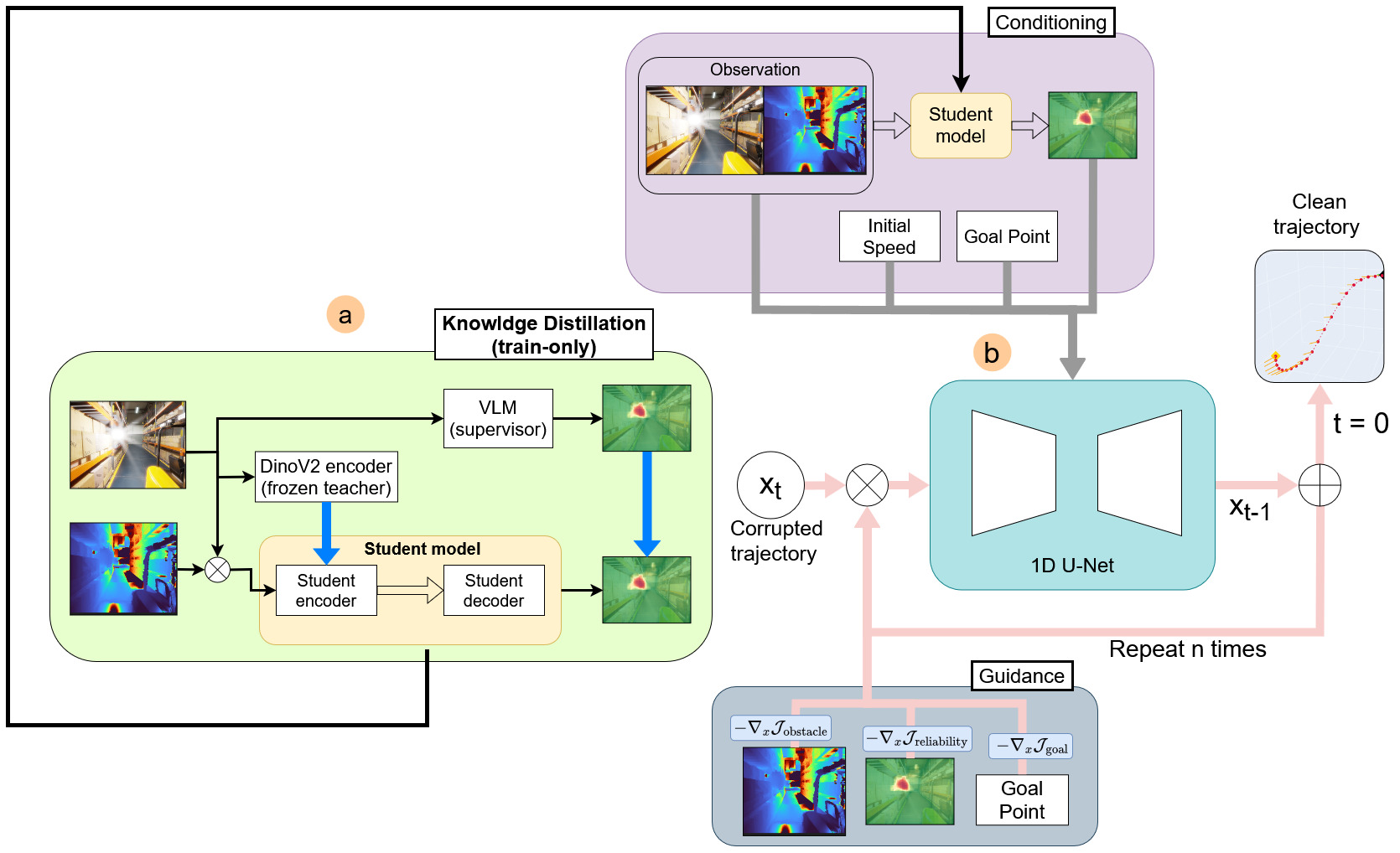}
  \caption{\textbf{Overview of the proposed system.}
    \textit{(a)}~Knowledge distillation pipeline for faster inference.
    \textit{(b)}~Diffusion-based trajectory planner.}
  \label{fig:system}
\end{figure}

Our system is a diffusion-based ego-trajectory planner with context of observation, goal point, intial speed for dynamics propagation and estimated reliability heatmap to avoid zones of unreliable perception. Its structure is presented in Figure~\ref{fig:system}

\subsection{Reliability estimation}
\label{subsec:REst}
Indoor RGBD sensors fail under overexposure, specular reflections, and
transparent surfaces, corrupting both RGB and depth channels simultaneously.
We condition the planner on a dense reliability map $M\in[0,1]^{H\times W}$,
where low values indicate perceptually unreliable regions.
The map is produced offline by AnyTraverse~\cite{sahu2025anytraverseA},
repurposed to score sensor reliability against indoor failure concepts
(Table~\ref{tab:visibility_weights}):
\begin{equation}
  M = 1 - \mathrm{AnyTraverse}(I;\,\mathbf{w}).
\end{equation}
 
\begin{table}[h]
\centering
\caption{Perceptual failure concept weights $\mathbf{w}$.}
\label{tab:visibility_weights}
\begin{tabular}{lc}
\toprule
\textbf{Concept} & $w_i$ \\
\midrule
Glass & 1.0 \\
Flare / overexposure             & 1.0 \\
Reflective surface                          & 1.0 \\
Glossy surface & 1.0 \\
\bottomrule
\end{tabular}
\end{table}
 
As AnyTraverse is too heavy for deployment,
safety maps are precomputed offline and used to supervise a lightweight
MiT-B0 student~\cite{xie2021segformer} with depth fusion at all encoder
scales. A frozen ViT-Adapter/DINOv2~\cite{oquab2024dinov2} backbone
provides multi-scale feature supervision $\mathcal{F}_T$, anchoring the
student's internal scene representation alongside the pixel-level targets
$M_T$ from AnyTraverse.

\paragraph*{Training objective.}
Per-level $1{\times}1$ adapters align student features to the teacher
dimension before computing feature imitation via cosine dissimilarity:
\begin{equation}
  \mathcal{L}_{\mathrm{feat}}
  = \frac{1}{|\mathcal{S}|}\sum_{s\in\mathcal{S}}
  \mathbb{E}\!\left[1 - \frac{\langle\hat{F}_S^{(s)},\,F_T^{(s)}\rangle}
                             {\|\hat{F}_S^{(s)}\|\cdot\|F_T^{(s)}\|}\right].
\end{equation}

Pixel-level regression uses hard-negative weighted L1, with unsafe pixels
receiving weight $w_i = 1+w_{max}(1-M_{T,i})$, $w_{max}{=}4$:
\begin{equation}
  \mathcal{L}_{\mathrm{task}}
  = \frac{\sum_i w_i\,|M_{S,i} - M_{T,i}|}{\sum_i w_i}.
\end{equation}
An SSIM term preserves the safe/unsafe boundary structure:
\begin{equation}
  \mathcal{L}_{\mathrm{SSIM}} = 1 - \mathrm{SSIM}(M_S,\,M_T).
\end{equation}
The total objective is:
\begin{equation}
\label{eq:distill_loss}
  \mathcal{L}_{\mathrm{distill}}
  = w_{\mathrm{feat}}\,\mathcal{L}_{\mathrm{feat}}
  + w_{\mathrm{task}}\bigl[
      (1-w_{\mathrm{ssim}})\,\mathcal{L}_{\mathrm{task}}
      + w_{\mathrm{ssim}}\,\mathcal{L}_{\mathrm{SSIM}}
    \bigr].
\end{equation}
\subsection{Diffusion-Based Trajectory Planner}
\label{sec:dif}

We adopt the DDPM formulation~\cite{ho2020ddpm}. The forward process
corrupts a clean trajectory $x_0$ over $T$ steps:
\begin{equation}{x_t = \sqrt{\bar{\alpha}_t}\, x_0 + \sqrt{1-\bar{\alpha}_t}\, \varepsilon, \qquad \varepsilon \sim \mathcal{N}(0,\mathbf{I})}\end{equation}
where $\bar{\alpha}_t = \prod_{s=1}^{t}(1-\beta_s)$.

A 1D U-Net $\varepsilon_\theta$ is trained to predict the added noise:
\begin{equation}
  \mathcal{L}_{\mathrm{diff}}
  = \mathbb{{E}_{t,x_0,\varepsilon}}
    \bigl[\|\varepsilon - \varepsilon_\theta(x_t,t,c)\|^2\bigr].
\end{equation}
The reverse process iteratively denoises from
$x_T\!\sim\!\mathcal{N}(0,\mathbf{I})$:
\begin{equation}
\label{eq:backward}
  x_{t-1} = \frac{1}{\sqrt{\alpha_t}}\!\left(x_t
    - \frac{\beta_t}{\sqrt{1-\bar{\alpha}_t}}\,
      \varepsilon_\theta(x_t,t,c)\right)
    + \sqrt{\tilde{\beta}_t}\,z,
  \quad z\sim\mathcal{N}(0,\mathbf{I}).
\end{equation}

\paragraph*{State representation.}
Two representations are compared: absolute ego-frame waypoints
$s_i\in\mathbb{R}^3$ and successive displacements
$\Delta s_i = s_{i+1}-s_i$.
Both are normalised channel-wise to $[-1,1]^3$ via min-max statistics
computed on the training set.
\paragraph*{Conditioning.}
The denoiser is conditioned on $c = \{v_0,\,g,\,F_I,\,F_S\}$.
Compact vectors — initial speed $v_0$ and goal $g$ — are injected
via FiLM. Spatially structured inputs are injected via two sequential
cross-attention blocks at the bottleneck: Image features from student encoder or VLM encoder $F_I$ provide scene-level context; reliability map
patch tokens $F_S$ provide reliability context.

\paragraph*{Auxiliary losses.}
Three auxiliary terms are added to the diffusion loss to enforce
trajectory constraints:
\begin{equation}
  \mathcal{L} = \mathcal{L}_{\mathrm{diff}}
    + \lambda_1\mathcal{L}_{\mathrm{smooth}}
    + \lambda_2\mathcal{L}_{\mathrm{origin}}
    + \lambda_3\mathcal{L}_{\mathrm{speed}}
    + \lambda_4\mathcal{L}_{\mathrm{safe}},
\end{equation}
where $\mathcal{L}_{\mathrm{smooth}}$ penalises squared second-order
finite differences (acceleration proxy); $\mathcal{L}_{\mathrm{origin}}$
anchors the first waypoint at the UAV origin (absolute representation
only); $\mathcal{L}_{\mathrm{speed}}$ enforces initial velocity
consistency; and $\mathcal{L}_{\mathrm{safe}}$ penalises waypoints
projecting onto low-reliability image regions:
\begin{equation}
  \mathcal{L}_{\mathrm{safe}}
  = \frac{1}{|\mathcal{V}|}\sum_{i\in\mathcal{V}}(1-\tau_i),
\end{equation}
where $\tau_i\in[0,1]$ is the sampled reliability score and
$\mathcal{V}$ is the set of in-frame waypoints.
\subsection{Guidance}
\label{ssec:guidance}

A trained diffusion model generates samples consistent with its training
distribution. To adapt at deployment without retraining, we apply
\emph{training-free} guidance~\cite{dhariwal2021diffusion}: the denoising
score is augmented with the gradient of a differentiable cost function,
\begin{equation}
  \tilde{\varepsilon}_\theta(x_t,t)
  = \varepsilon_\theta(x_t,t,c)
  - s\,\nabla_{x_t}\mathcal{C},
\end{equation}
where $s$ is the guidance scale. The total cost decomposes as:
\begin{equation}
  \mathcal{C} = w_{\mathrm{obs}}\,\mathcal{C}_{\mathrm{obs}}
              + w_{\mathrm{goal}}\,\mathcal{C}_{\mathrm{goal}}
              + w_{\mathrm{perc}}\,\mathcal{C}_{\mathrm{perc}}.
\end{equation}

\paragraph*{Obstacle cost.}
$\mathcal{C}_{\mathrm{obs}}$ is the CHOMP cost ~\cite{ratliff2009chomp}
evaluated over a 3-D ESDF built from the depth point cloud.
For each waypoint with signed distance $d_i$ to the nearest obstacle:
\paragraph*{Goal cost.}
$\mathcal{C}_{\mathrm{goal}}$ penalizes the distance from the final
waypoint to the target position $\mathbf{g}\in\mathbb{R}^3$:
\begin{equation}
  \mathcal{C}_{\mathrm{goal}} = \|\mathbf{p}_H - \mathbf{g}\|_2^2.
\end{equation}

\paragraph*{Reliability cost.}
Pixels are split into two reliability stages using thresholds
$\tau_h < \tau_s$ and $\tau_h = 0.3$, $\tau_s = 0.6$ in our implementation.. Pixels with $M_S \leq \tau_h$ are back-projected
along rays at different depths and concatenated with the
depth point cloud before ESDF construction, making them hard obstacles
absorbed into $\mathcal{C}_{\mathrm{obs}}$.
Pixels with $\tau_h < M_S \leq \tau_s$ populate a separate weaker CHOMP cost
with activation distance $\eta_{\mathrm{perc}} > \eta_{\mathrm{obs}}$, as was done for obstacle avoidance.

\paragraph*{Guidance step: absolute representation.}
At each step $t$ the noisy trajectory is projected onto the estimated
clean manifold~\cite{wang2024optimizing}:
\begin{equation}
  \hat{x}_0 = \frac{x_t - \sqrt{1-\bar{\alpha}_t}\,
                    \varepsilon_\theta(x_t,t,c)}
                   {\sqrt{\bar{\alpha}_t}}.
\end{equation}
The cost gradient is computed in metric coordinates, scaled by
anisotropy $A=[a_x,a_y,a_z]$.
A new noise estimate is derived from the updated $\hat{x}_0^*$
in DDIM style~\cite{song2020ddim}:
\begin{equation}
\label{eq:eff_noise}
  \hat{\varepsilon}
  = \frac{x_t - \sqrt{\bar{\alpha}_t}\,\hat{x}_0^*}
         {\sqrt{1-\bar{\alpha}_t}}.
\end{equation}

The reverse step ~\eqref{eq:backward} is then applied with
$\hat{\varepsilon}$ in place of $\varepsilon_\theta$ to obtain
$x_{t-1}$.

\paragraph*{Guidance step: delta representation.}
Waypoint positions are recovered by cumulative summation
$\mathbf{p}_k = \sum_{i=0}^{k}\Delta s_i$.
The position gradient is converted to delta space via the transpose
of the cumsum operator and applied directly to the denoised state.

Unlike the delta variant, model reconstructs the full structure of the trajectory. In this case shifts of guidance destroy inner manifold of the diffusion and resulted trajectory becomes inappropriate.

\section{Dataset}
\begin{figure}[t]
    \centering
    \includegraphics[width=0.65\textwidth]{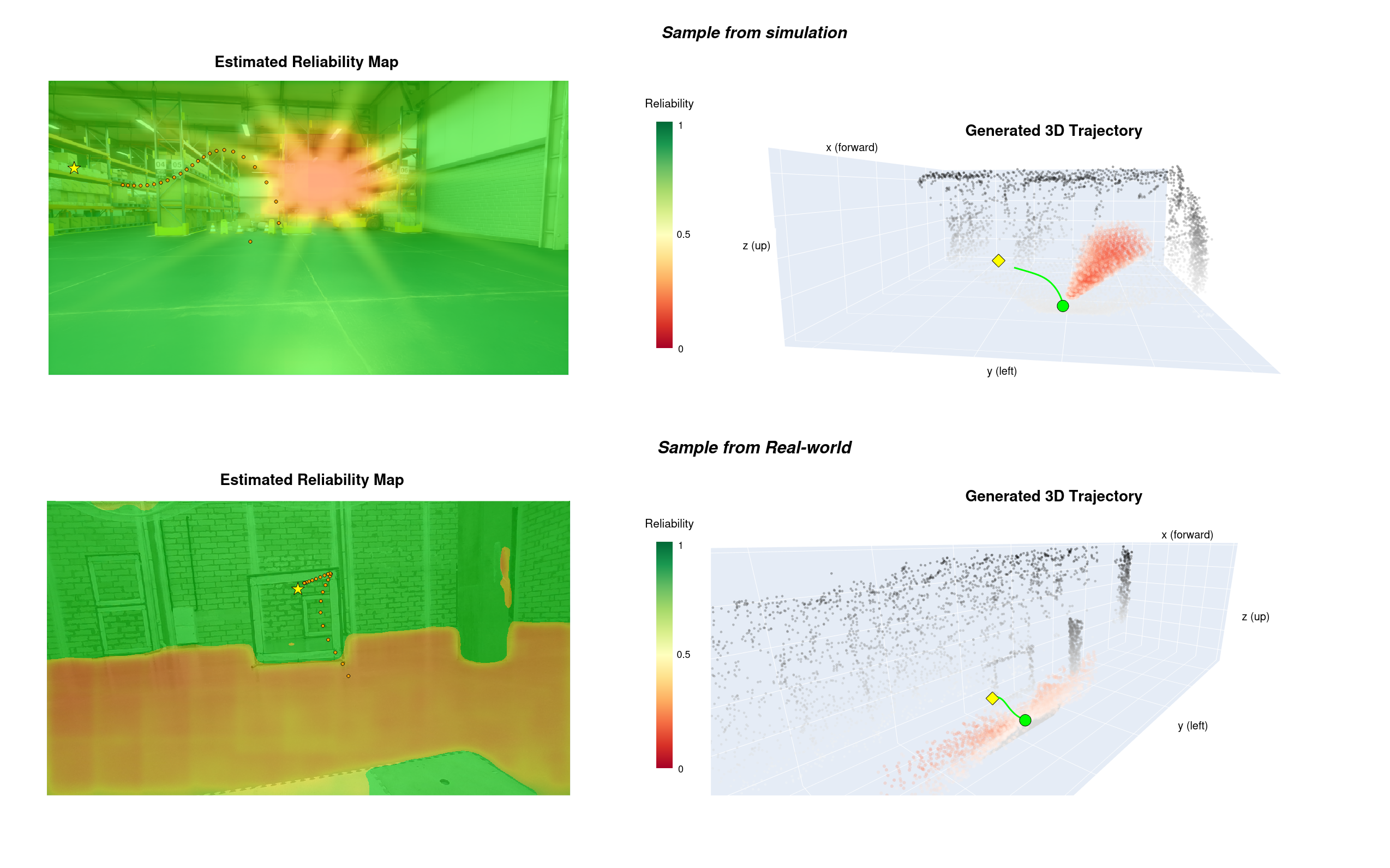}
    \caption{\textbf{Training dataset examples} for simulation (top) and the real world (bottom).
\textit{Left:} the RGB-estimated reliability map conditioning the diffusion model (green
reliable, red unreliable), flagging glare in the warehouse and the reflective floor in the
real scene. \textit{Right:} the target trajectory (green) from start (yellow diamond) to goal
(green circle), with the scene point cloud (grey) and low-reliability regions as virtual
obstacles (red).}
    \label{fig:dataset_example}
\end{figure}

We generate a synthetic dataset of quadrotor trajectories in warehouse environments
simulated in Pegasus Simulator~\cite{10556959}. For each scene,
a four-stage pipeline performs stereo depth reconstruction, estimates a perceptual
reliability map, plans a geometric path, and simulates the corresponding physics-based
trajectory, producing the paired observation, reliability map, and target trajectory shown
in Fig.~\ref{fig:dataset_example}.
\section{Real-World Deployment}
We evaluate our system on a custom quadrotor built around an 8-inch frame. Low-level flight
control is handled by ArduPilot, with high-level commands issued through MAVROS. Visual observations are captured
with an Intel RealSense D455 depth camera, while trajectory generation is performed off-board on a single NVIDIA RTX 4070 GPU with the memory usage of $500$\,MB.
Replacing the vision-language model with the distilled student network allows the full
framework to run up to $2\times$ faster, bringing reliability estimation within the real-time
planning budget.
\label{sec:result}

Our evaluation is organized around three questions:

\begin{enumerate}
    \item Does conditioning trajectory generation on perceptual reliability improve
    navigation safety relative to a state-of-the-art diffusion planner that relies on
    visual conditioning alone?
    \item How does guided sampling affect the feasibility and quality of the generated
    trajectories?
    \item What are the individual and combined contributions of reliability conditioning
    and guidance?
\end{enumerate}

To answer the first question, we compare against NoMaD~\cite{sridhar2023nomad}, a strong
diffusion-based baseline that, like our method, generates trajectories from visual
observations, but has no representation of sensor reliability and therefore cannot
distinguish trustworthy regions from perceptually degraded ones. This isolates the effect of
reliability awareness while holding the generative formulation fixed. To answer the second
and third questions, we ablate our method across the four combinations of reliability
conditioning (on or off) and ESDF guidance (on or off), measuring the effect of each
component in isolation and their interaction. This design separates the contribution of
\emph{what} the planner is told about the scene from \emph{how} the denoising process is
steered.
\subsection{Metrics}

Path efficiency represents how straight the trajectory is:
$$\text{Path Efficiency} = \frac{\text{straight-line distance (start → end)}}{\text{actual path length}}$$
Value of 1.0 = perfectly straight line. Less than 1 = detours.

Safety is characterized by two main parameters:
\begin{itemize}
    \item Violation. Ratio of collided waypoints of generated trajectory to the whole sample.
    \item Minimal Distance to an Obstacle. Minimal Signed Distance Function(SDF) of waypoints.
\end{itemize}

The reliability map $\mathcal{R}: \Omega \rightarrow [0, 1]$ is predicted
by the student model for each image pixel. Each waypoint $\mathbf{p}_t$
is projected onto the image plane.

\subsection{Ablation Study}
Table~\ref{tab:results1} compares the two trajectory representations under varying
guidance. With guidance disabled, both representations yield short and efficient paths,
but neither avoids obstacles or unreliable regions. Once guidance is enabled, the two
diverge sharply: the ego-path representation retains nearly double the path efficiency
($0.590$ vs.\ $0.288$) and ends almost three times closer to the goal ($0.710$\,m vs.\
$1.959$\,m) than the delta-path representation, whose relative increments accumulate error
under the guidance gradients and drift away from the goal. We therefore adopt the ego-path
representation. The increase in trajectory length and reduction in nominal efficiency under
guidance reflect the detours required to route around physical obstacles and low-reliability
regions, and should be read together with the safety metrics in Table~\ref{tab:results}
rather than in isolation.
\begin{table}[h]
\centering
\caption{Ablation Study}
\label{tab:results1}
\begin{tabular}{lccc}
\toprule
\textbf{Method}
  & \thead{Trajectory \\ length, m ($\downarrow$)}
  & \thead{Path \\ efficiency($\uparrow$)}
  & \thead{Mean distance \\ to goal point, m ($\downarrow$)} \\
\midrule
Ours (ego-path) 
  & $5.027 \pm 0.871$
  & $0.590 \pm 0.082$
  & $0.710 \pm 0.317$ \\
Ours (delta-path)
  & $ 4.724\pm 0.431$
  & $0.288 \pm 0.105$
  & $1.959 \pm 0.472$\\
\midrule
\makecell[l]{Ours (ego-path)\\ w/o Reliability}
  & $5.180 \pm 0.917$
  & $0.564 \pm 0.079 $
  & $0.523 \pm 0.245$ \\
\makecell[l]{Ours (delta-path)\\ w/o Reliability}
  & $4.654 \pm 0.467$
  & $ 0.297 \pm 0.098$
  & $1.876 \pm 0.497$ \\
\midrule
\makecell[l]{Ours (ego-path) w/o Guidance}
  & $2.869 \pm 0.485$
  & $0.867 \pm 0.014$
  & $0.293 \pm 0.183$ \\
\makecell[l]{Ours (delta-path) w/o Guidance}
  & $\mathbf{2.664 \pm 0.407}$
  & $\mathbf{0.991 \pm 0.009}$
  & $\mathbf{0.228 \pm 0.115 }$ \\
\bottomrule
\end{tabular}
\end{table}

Adding blocks of guidance steers trajectory to avoid obstacles and reliable zones for safer navigation rather than direct goal reaching.

\subsection{Comparison with a Baseline}
Table~\ref{tab:results} reports the safety and perceptual-quality metrics. Our planner
substantially outperforms NoMaD~\cite{sridhar2023nomad}, a visually-conditioned diffusion
baseline without any notion of reliability, across every metric: it increases the minimum
distance to obstacles by almost two orders of magnitude ($0.009$\,m to $0.562$\,m), reduces
the obstacle-violation rate from $40.3\%$ to below $10\%$, and raises the mean reliability of
traversed regions from $0.588$ to $0.925$. The reliability-guidance term is the dominant
driver of this perceptual safety: removing it collapses the mean reliability of the ego-path
variant from $0.898$ to $0.293$, confirming that the term actively steers trajectories into
trustworthy regions rather than merely correlating with them. Although the delta-path variant
attains marginally better raw safety scores, it fails to reach the goal reliably
(Table~\ref{tab:results1}); the ego-path representation provides the best balance of safety,
high clearance ($0.723$\,m), and goal-directed behaviour, and is our final configuration.

\begin{table}[h]
\centering
\caption{Quantitative comparison of planning methods.
         Best results are highlighted in \textbf{bold}.}
\label{tab:results}
\begin{tabular}{lcccc}
\toprule
\textbf{Method}
  & \thead{Min. Distance\\to an Obstacle, m  ($\uparrow$)}
  & \thead{Clearance, m \\ ($\uparrow$)}
  & \thead{Obstacle \\ Violation, \%  ($\downarrow$)}
  & \thead{Mean \\ Reliability ($\uparrow$)} \\
\midrule
NoMaD~\cite{sridhar2023nomad}
  & $0.009 \pm 0.126$
  & $0.154 \pm 0.193$
  & $40.3 \pm 10.283$
  & $0.588 \pm 0.145$ \\
\midrule
Ours (ego-path)
  & $0.532 \pm 0.259$
  & $\mathbf{0.723 \pm 0.188}$
  & $9.6 \pm 17.4$
  & $0.898 \pm 0.194$ \\
Ours (delta-path)
  & $\mathbf{ 0.562 \pm 0.227}$
  & $0.588 \pm 0.145$
  & $\mathbf{9.5 \pm 8.0}$
  & $\mathbf{0.925 \pm 0.120}$ \\
\midrule
\makecell[l]{Ours (ego-path) \\ w/o Reliability}
  & $0.368 \pm 0.255$
  & $0.615 \pm 0.147$
  & $9.0 \pm 0.133$
  & $0.783 \pm 0.183$ \\
\makecell[l]{Ours (delta-path) \\ w/o Reliability}
  & $0.519 \pm 0.207$
  & $0.434\pm 0.099$
  & $10.6 \pm 9.6$
  & $0.912 \pm 0.204$ \\
\bottomrule
\end{tabular}
\end{table}

Guidance make generator safer in terms of obstacle avoidance and steering to reliable regions. Usage only of visual scene understanding is not enough to navigate in suddenly degrated unknown area.


\section{Conclusion and Discussion}
We presented a reliability-aware diffusion planner for autonomous 3D UAV navigation in
unstructured indoor environments. End-to-end generative planners trained on clean
observations lack any explicit mechanism to recognise when perception is unreliable and
treat degraded regions as valid evidence for planning. Our central idea is to make sensing
reliability a first-class input to trajectory generation. To this end we (i) formulate
navigation as an end-to-end diffusion policy mapping raw observations to 3D trajectory
distributions, (ii) condition generation on a scene-level reliability heatmap that flags
regions where depth cannot be trusted, (iii) distil this open-vocabulary reasoning from a
vision-language model into a lightweight network that runs within the planning budget, and
(iv) steer sampling with a training-free two-stage ESDF cost over physical and reliability-induced
virtual obstacles.

Reliability awareness is the decisive contribution. Against NoMaD~\cite{sridhar2023nomad}, a
visually-conditioned diffusion baseline without it, our planner raises the minimum distance to
obstacles from $0.009$\,m to $0.562$\,m, lowers the obstacle-violation rate from $40.3\%$ to
$9.6\%$, and increases the mean reliability of traversed regions from $0.588$ to $0.925$.
Ablating the reliability-guidance term alone collapses mean reliability from $0.898$ to
$0.783$, confirming that it actively routes trajectories through trustworthy regions rather
than merely correlating with safer paths.

The approach has limitations. The reliability heatmap is bounded by the teacher model and the
residual distillation gap, remains scene-level rather than pixel-calibrated, and the per-step
guidance cost scales with the number of denoising steps and ESDF queries. These motivate two
directions for future work: generalisation targeting, extending the reliability estimator to
outdoor, dynamic, and adversarially lit conditions with domain adaptation that narrows the
teacher-student gap and supports cross-embodiment transfer; and further onboard optimisation
of guidance through amortised or learned guidance, fewer-step samplers, and efficient on-device
ESDF queries to raise the achievable replanning rate on embedded hardware.




\bibliography{example}  

\end{document}